\documentclass[english]{IEEEtran}
\usepackage[T1]{fontenc}
\usepackage[latin9]{inputenc}
\usepackage{graphicx}

\makeatletter
\@ifundefined{showcaptionsetup}{}{%
 \PassOptionsToPackage{caption=false}{subfig}}
\usepackage{subfig}
\makeatother

\usepackage{babel}
\begin{document}
\title{Collaborative Active Learning in Conditional Trust Environment}
\author{%
\begin{minipage}[t]{0.3\textwidth}%
\begin{center}
Zan-Kai Chong\\
\emph{School of Science and\\ Technology\\
Kwansei Gakuin University}\\
Japan\\
zankai@ieee.org
\par\end{center}%
\end{minipage}%
\begin{minipage}[t]{0.3\textwidth}%
\begin{center}
Hiroyuki Ohsaki\\
\emph{School of Science and\\ Technology\\
Kwansei Gakuin University}\\
Japan\\
ohsaki@kwansei.ac.jp
\par\end{center}%
\end{minipage}%
\begin{minipage}[t]{0.3\textwidth}%
\begin{center}
Bryan Ng\\
\emph{ 
School of Engineering \& Computer Science\\
Victoria University of Wellington}\\
New Zealand
\\
bryan.ng@ecs.vuw.ac.nz
\par\end{center}%
\end{minipage}}
\maketitle
\begin{abstract}
In this paper, we investigate collaborative active learning, a paradigm
in which multiple collaborators explore a new domain by leveraging
their combined machine learning capabilities without disclosing their
existing data and models. Instead, the collaborators share prediction
results from the new domain and newly acquired labels. This collaboration
offers several advantages: (a) it addresses privacy and security concerns
by eliminating the need for direct model and data disclosure; (b)
it enables the use of different data sources and insights without
direct data exchange; and (c) it promotes cost-effectiveness and resource
efficiency through shared labeling costs. To realize these benefits,
we introduce a collaborative active learning framework designed to
fulfill the aforementioned objectives. We validate the effectiveness
of the proposed framework through simulations. The results demonstrate
that collaboration leads to higher AUC scores compared to independent
efforts, highlighting the framework's ability to overcome the limitations
of individual models. These findings support the use of collaborative
approaches in active learning, emphasizing their potential to enhance
outcomes through collective expertise and shared resources. Our work
provides a foundation for further research on collaborative active
learning and its practical applications in various domains where data
privacy, cost efficiency, and model performance are critical considerations.
\end{abstract}

\section{Introduction}

In today's rapidly evolving market landscape, collaborative approaches
in machine learning are becoming increasingly vital for businesses
seeking to stay ahead of the curve. A prime example of this is the
partnership between Apple and Goldman Sachs in creating the Apple
Card, tailored for iPhone users \cite{goldmansachs2019apple}. Another
instance is observed in Malaysia, where traditional financial institutions
and tech companies are forming alliances in digital banking \cite{kapron2021malaysian}.

These ventures combine technological and financial services expertise,
suggesting collaborative effort to leverage the strengths of each
collaborator in informed decision-making, such as developing more
accurate and cost-effective machine learning models. Although unreported
in the news, we can plausibly infer that these partnerships are likely
based on \emph{conditional trust} that prevents them from disclosing
or sharing proprietary models or sensitive data, as to adhere to the
confidentiality and data privacy regulations.

A few key concepts need to be introduced here to address collaboration
in a conditional trust environment. Generally, learning a new domain
is challenging due to the labelling cost and the associated risks.
Therefore, one can explore a new domain with active learning as it
enables rapid and cost-effective exploration by selectively labeling
instances that are most beneficial to machine learning models. Many
active learning sampling strategies have been studied by the research
community, such as uncertainty sampling \cite{lewis1994heterogeneous},
query-by-committee \cite{freund1997selective}, bell-curve sampling
\cite{Chong2023}, etc. A comprehensive survey can be found at \cite{tharwat2023survey}.
When multiple parties work together towards the same active learning
objective, they are engaged in collaborative active learning.

On the other hand, the notion of machine learning with no data sharing
has been studied in federated learning (FL) \cite{GoogleFederatedLearning}.
FL is a form of collaborative machine learning that allows multiple
devices to train a model using their local data. These local updates
are then sent to a central server, which aggregates them to enhance
the global model. Subsequently, the updated model is redistributed
back to the clients. Despite the widespread studies on FL \cite{wang2023unlocking,wen2023survey},
its applicability to our research scenario is limited as both existing
data and models cannot be shared among collaborators.

Another similar learning method is called swarm learning (SL) \cite{warnat2021swarm}.
SL is a form of cooperative machine learning that involves the sharing
of model parameters. In this approach, individual nodes train a model
on their local data and then share the model parameters with the rest
of the swarm using block chain technology. These parameters are then
merged to improve the overall model. SL has gained the attention of
the medical research community \cite{han2022demystifying,saldanha2022swarm}.
However, this learning method is not suitable for our scenario as
we do not presuppose the use of a uniform learning algorithm across
all collaborators. Hence, sharing the model parameters are not universally
applicable.

Another interesting approach such as \cite{li2022can} uses Model-Sharing
Strategy (MSS) to address the challenges of sharing sensitive data.
The authors propose MSS as a means to leverage the potential of closed
data, which are not readily shareable due to privacy, ownership, trust,
and incentive issues. Instead of sharing the data, the other party
will send the model to another collaborator to train and return back
the model. This approach allows data to remain invisible but interoperable
through models, significantly reducing the risk of data leakage compared
to conventional data sharing. Again such approach is not applicable
in our research scenario.

In this paper, we explore a collaborative active learning framework
in a conditional trust environment, where collaborators do not share
their data and models. We introduce the term Conditionally Collaborative
Active Learning (C2AL) to succinctly refer to this problem setting.
C2AL offers several advantages while imposing certain considerations:
\begin{description}
\item [{(a)}] Privacy and security: C2AL eliminates the need for direct
model and data disclosure, addressing critical privacy and security
considerations.
\item [{(b)}] Diverse data utilization: C2AL allows leveraging different
data sources and insights without the need for direct data exchange.
\item [{(c)}] Cost and resource efficiency: C2AL promotes cost-effectiveness
and resource efficiency by enabling shared labeling costs.
\end{description}
To address the constraints of C2AL, we propose a collaborative framework
that enables multiple collaborators to jointly explore a new domain
using active learning while prioritizing data and model privacy. In
this framework, collaborators refrain from disclosing their existing
data and models to each other. Instead, they share prediction results
on unlabeled instances and jointly select new instances for labeling.
The acquired labels are then shared among all collaborators and used
to train improved predictive ensemble models that leverage the collective
knowledge captured in the shared prediction results.

The remainder of this paper is organized as follows: Section \ref{sec:Collaborative-Framework}
discusses the proposed collaborative framework, including the algorithm
and its theoretical foundations. Section \ref{sec:Simulation} presents
the simulation results, demonstrating that multiple collaborators
can leverage each other's model prediction capabilities to achieve
better performance compared to a collaborator learning independently.
Finally, Section \ref{sec:Conclusion} draws the conclusion.

\section{Collaborative Framework \label{sec:Collaborative-Framework}}

In this section, we first present the outline of the collaborative
framework for active learning, followed by a detailed discussion of
the steps involved.

\subsection{Outline of Collaborative Framework \label{subsec:framework-outline}}

Our proposed collaborative active learning framework involves two
or more collaborators, one of which serves as the coordinator responsible
for acquiring and disseminating labels. In general, this collaboration
operates under the following constraints:
\begin{enumerate}
\item No sharing of existing data, and
\item No disclosure of existing models.
\end{enumerate}
In addition, we also make the following assumptions:
\begin{enumerate}
\item Collaborators adhere to the collaboration framework.
\item In addition to the basic data, each collaborator may possess unique
pieces of information about the instances, known only to that individual
collaborator, providing them with distinct insights and perspectives.
\label{enu:assumptions-data}
\item Each collaborator may prefer a different learning algorithm. \label{enu:assumption-learning-algo}
\item Each collaborator possesses an initial base model, the performance
of which is unknown in a new domain.
\end{enumerate}
Specifically, Assumptions \ref{enu:assumptions-data} and \ref{enu:assumption-learning-algo}
are the primary reasons for the varying predictive power among the
machine learning models that developed by different collaborators.

\begin{figure*}
\begin{centering}
\subfloat[]{\begin{centering}
\includegraphics[width=0.3\textwidth]{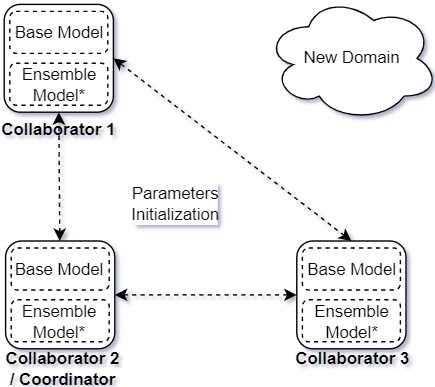}
\par\end{centering}
}\subfloat[]{\begin{centering}
\includegraphics[width=0.3\textwidth]{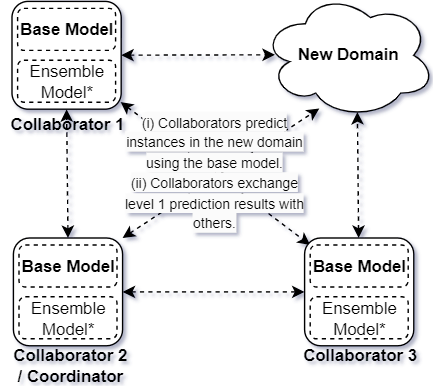}
\par\end{centering}
}\subfloat[]{\begin{centering}
\includegraphics[width=0.3\textwidth]{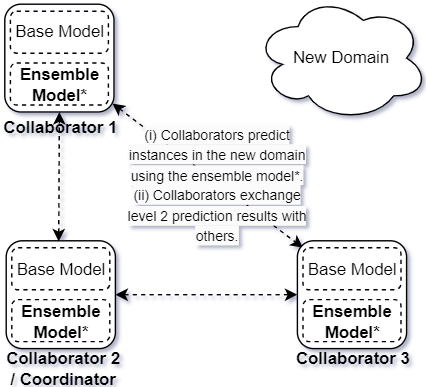}
\par\end{centering}
}
\par\end{centering}
\begin{centering}
\subfloat[]{\begin{centering}
\includegraphics[width=0.3\textwidth]{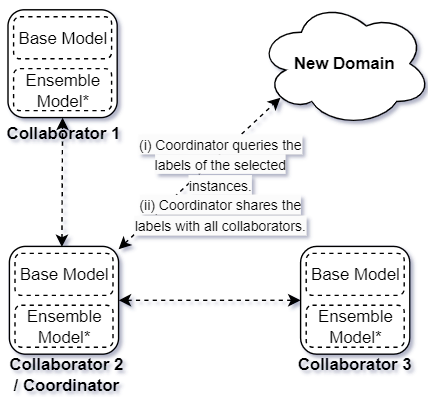}
\par\end{centering}
} \ \ \ \ \ \ \ \ \ \ \subfloat[]{\begin{centering}
\includegraphics[width=0.3\textwidth]{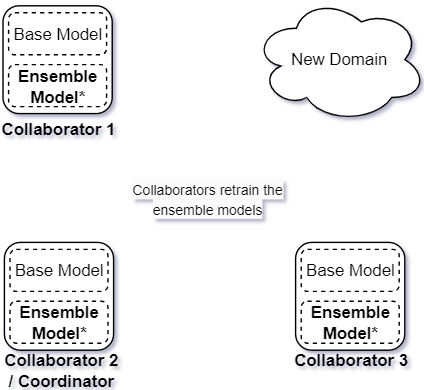}
\par\end{centering}
}
\par\end{centering}
\caption{Collaborative process among three collaborators: (a) Parameter initialization
and coordinator appointment. (b) Collaborators use their individual
base models to predict instances in the new domain and share the initial
results. (c) Leveraging the shared results, collaborators employ ensemble
models for refined predictions and disseminate updated results. (d)
The coordinator acquires labels and distributes them to all collaborators.
(e) Collaborators update the ensemble models with the newly acquired
labels and initiate the next query cycle starting from step (b). \label{fig:Collaborative-process}}
\end{figure*}

As illustrated in Figure \ref{fig:Collaborative-process}, the collaborative
active learning process proceeds as follows within the context outlined
above:
\begin{itemize}
\item Step 1 - Parameters Initialization: Collaborators engage in a preparatory
phase to outline the parameters of the collaboration. Key parameters
encompass:
\begin{enumerate}
\item Determining the total rounds of query, represented as $q$, and the
total collaborators as $k$.
\item Specifying the number of instances to query in each round, represented
as $n$.
\item The pre-agreed sampling function, e.g., uncertainty sampling.
\item Appointing a coordinator. 
\end{enumerate}
\item Step 2 - Exchange of Level 1 Prediction Results: All collaborators
evaluate the available unlabelled instances using their existing base
models. Then, the prediction results are shared with the others.
\item Step 3 - Exchange of Level 2 Prediction Results: Upon receiving Level
1 prediction results, each collaborator generates Level 2 prediction
results using the latest ensemble model and subsequently shares these
with all participants, including the coordinator. It should be noted
that Level 1 prediction results are part of the inputs for the ensemble
model. In cases where a collaborator has not yet developed the ensemble
model (e.g., due to insufficient new labels), the original Level 1
prediction results are circulated as the Level 2 prediction results.
\item Step 4 - Instance Acquisition: After receiving Level 2 prediction
results, the coordinator determine the best instances to query using
the algorithm illustrated in Section \ref{subsec:Collaborative-Instance-Selection}.
Subsequently, the new labels are acquired and disseminated to all
collaborators.
\item Step 5: Enhancing Model Predictive Power and Iteration: In this step,
a collaborator can begin to enhance its predictive power by (re-)building
an ensemble model that leverages all the received Level 1 prediction
results and new labels, provided that a sufficient number of them
have been accumulated. Subsequently, Steps 1 through 4 are repeated
for a total of $q-1$ rounds.
\end{itemize}

\subsection{Instance Selection and Acquisition \label{subsec:Collaborative-Instance-Selection}}

We further elaborate on the instance selection process outlined in
Steps 2 to 4 here. For simplicity, we will use \textquotedbl A\textquotedbl{}
to denote the first collaborator and \textquotedbl B\textquotedbl{}
to refer to the second collaborator or, when applicable, the rest
of collaborators.

Recall that all collaborators have access to different details of
all unlabeled instances. In Step 2, collaborators evaluate the unlabeled
instances using their base models and exchange the prediction results
(Level 1). They then incorporate received prediction results into
their ensemble models (if applicable) and update the latest results
(Level 2) to the coordinator. If a collaborator does not have an ensemble
model, their Level 2 prediction results will remain unchanged from
Level 1.

In Step 4, suppose the coordinator receives two Level 2 prediction
results from collaborators A and B. At the absence of test data, the
coordinator has no way to evaluate the accuracy of the prediction
results shared by each collaborator. To minimize the risk, the coordinator
employ the pre-agreed sampling function to select the best instance
to query from each received Level 2 prediction result alternately.
This ensures that each collaborator has an equal opportunity to contribute
to the selection process. If an instance from either Level 2 prediction
result has already been selected, the coordinator moves to the next
instance in sequence. This process continues until a total of $n$
unique instances have been selected.

\subsection{Retrain Models \label{subsec:Retrain-Models}}

We provide an in-depth discussion on model retraining process delineated
in Step 5 of Section \ref{subsec:framework-outline}.

Recall that the collaboration is encapsulated in the process of collectively
developing models with collaborators, while preserving the privacy
of individual base models and existing data. To start the discussion,
we posit the existence of an optimal learning algorithm within the
active learning framework, underpinned by two principal assumptions:
\begin{enumerate}
\item Monotonic Improvement with Additional Labels: The performance of a
machine learning model will not deteriorate with the introduction
of carefully selected additional labels. Mathematically, let $P(M,L)$
denote the performance of model $M$ with a set of labelled instances
$L$, and let $L'$ denote an expanded set of labels such that $L'\supseteq L$,
then the model's performance with the expanded set is at least as
good as before, i.e., $P(M,L')\geq P(M,L)$.\label{enu:improvement-additional-label}
\item Monotonic Improvement in Ensemble Learning: We posit that in ensemble
learning, where learning strategies are dynamically adapted based
on the base learners, the performance of a system is expected to either
maintain or improve. Let $M_{1},\dots,M_{k}$ represent $k$ unique
base learners, and $F(M_{1},\dots M_{K})$ denote an ensemble learning
framework that dynamically selects or combines the base learners,
then the performance of this framework, denoted as $P(F(M_{1},\dots,M_{k}),L)$
for a set of labels $L$, is expected to be at least as good as the
better-performing strategy among $M_{1},\dots,M_{k}$, i.e., $P(F(M_{1},\dots M_{k}),L)\geq\max(P(M_{1},L),\dots,P(M_{k},L))$.
A detailed theoretical framework can be referenced in \cite{dietterich2000ensemble}
and recent developments are discussed in \cite{akano2022assessment,naderalvojoud2023improving,li2020comparative}.
\label{enu:improvement-ensemble-learning}
\end{enumerate}

Given the aforementioned assumptions, the collaborators will begin
building the ensemble models whenever they accumulate sufficient labels
and prediction results (as the input to ensemble models) from other
collaborators.

\section{Simulation\label{sec:Simulation}}

This section outlines the simulation setup and then discusses the
results. In particular, we demonstrate that the performance of collaborative
active learning outperforms that of a single collaborator learning
independently in C2AL environment.

\subsection{Simulation Setup}

We use Scikit-learn's \emph{make\_classification} function \cite{scikit-learn}
to create balanced synthetic binary datasets of desired characteristics.
These datasets contain equal instances of both classes, preventing
bias during training and evaluation. Each dataset comprises 3000 instances
with 20 features. We set a class separation factor of 0.7, indicating
moderate overlap between feature distributions of the two classes.
This overlap presents a more challenging classification task compared
to datasets with higher separation factors.

We employ the Area Under the Curve (AUC) score as the model performance
metric in our simulations, and adopt uncertainty sampling as the active
learning strategy. The interactions among collaborators are structured
according to the steps illustrated in Section \ref{sec:Collaborative-Framework}.
Base models are trained to achieve predefined AUC scores using a random
selection of 2 features and 100 instances. If the models fall short
of the AUC thresholds, we retrain them iteratively until the criteria
are satisfied. Upon reaching the target AUC scores, the remaining
dataset is classified as unknown data for the upcoming active learning
queries.

As outlined in Section \ref{subsec:framework-outline} and specified
in Assumption \ref{enu:assumptions-data}, each collaborator potentially
possesses distinct information about the unlabeled instances. This
distinction is achieved by granting each collaborator exclusive access
to unique features within the unlabeled data.

In active learning, we posit that collaborators are motivated to explore
the new domain by selecting most beneficial instances. In theory,
the collaborators have no access to test data, making the direct evaluation
of newly developed models unfeasible for them. However, as observers,
we are able to gauge the model performance by utilizing the test data.

\subsection{Simulation Results}

We present two simulations as follows. The first involves a single
collaborator (denoted as Collaborator 1) learning the new domain independently,
and the second comprises four collaborators, denoted as Collaborators
1 through 4, engaging in collaborative active learning.

\begin{figure}
\subfloat[]{
\begin{centering}
\includegraphics[width=0.95\columnwidth]{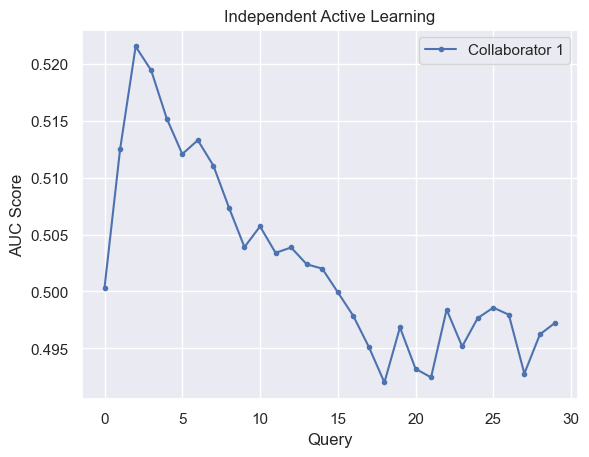}
\par\end{centering}

}

\subfloat[]{
\begin{centering}
\includegraphics[width=0.95\columnwidth]{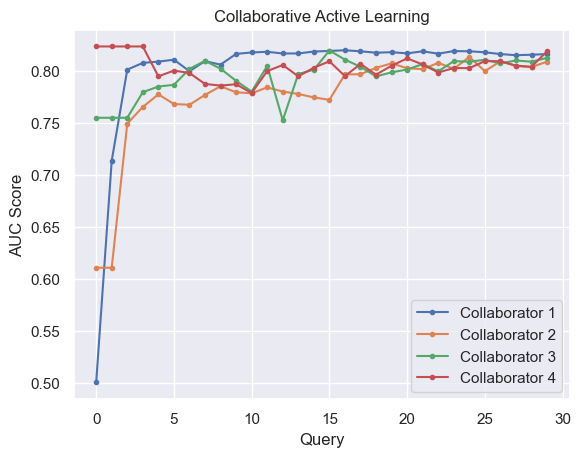}
\par\end{centering}

}

\caption{Progression of AUC scores with each additional query in active learning
for (a) a collaborator learning independently, and (b) four collaborators
engaging in collaborative active learning. \label{fig:auc-score}}
\end{figure}

Figure \ref{fig:auc-score}(a) presents the AUC score of Collaborator
1, who employs active learning independently throughout successive
queries in a simulation cycle. The base model is trained with a linear
algorithm to achieve a predetermined AUC score of 0.50 - 0.59. After
the first query, Collaborator 1 initiates the construction of an ensemble
model that integrates the base model with newly acquired labels. Due
to the inadequacy of the linear model in effectively segmenting classes
within a noisy and overlap dataset, we observed that Collaborator
1 is unable to query useful instances using uncertainty sampling.
Consequently, the subsequent ensemble model fails to show progress
throughout the simulation.

On the other hand, we simulate collaborative active learning among
four collaborators, with results presented in Figure 1(b). To test
the simulation with base models of different performance levels, the
base models of collaborators 1 through 4 are intentionally trained
to achieve AUC scores of 0.50 - 0.59, 0.60 - 0.69, 0.70 - 0.79, and
0.80 - 0.89, respectively. Following this, the collaborators commence
the construction of ensemble models at first, second, third, and fourth
queries. Collaborator 1 employs a linear algorithm, while the others
utilize random forests, gradient boosting machine and XGBoost respectively.

\begin{figure}
\subfloat[]{\begin{centering}
\includegraphics[width=0.95\columnwidth]{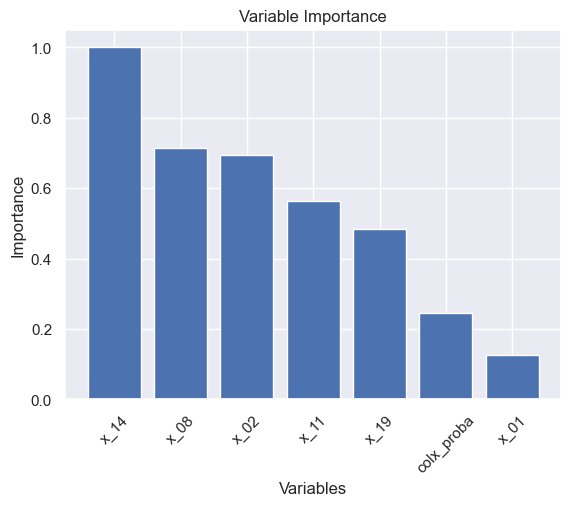}
\par\end{centering}
}

\subfloat[]{\begin{centering}
\includegraphics[width=0.95\columnwidth]{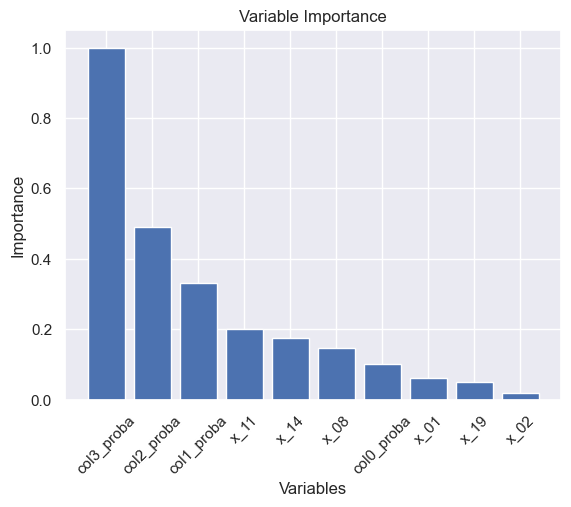}
\par\end{centering}
}

\caption{Variable importance plot for (a) a collaborator learning independently
in active learning, and (b) four collaborators engaging in collaborative
active learning. \label{fig:varimp}}
\end{figure}

Our observations suggest that all collaborators can swiftly leverage
the prediction results shared by others. By incorporating these insights
into new ensemble models, they consistently achieve high AUC scores,
ranging from approximately 0.80 to 0.85. For example, Figure \ref{fig:varimp}
illustrates the feature importances of Collaborator 1 when learning
independently (Figure \ref{fig:varimp} (a)) and when learning in
a collaborative framework (Figure \ref{fig:varimp} (b)). The variables
on the x-axis indicate the features accessible by Collaborator 1 (e.g.,
\textquotedbl x\_01\textquotedbl ) and the prediction results from
other collaborators (e.g., \textquotedbl col1\_proba\textquotedbl ).
Despite the initial poor performance of the base model, Collaborator
1 has effectively utilized the shared prediction results from others
and swiftly integrated them into its new ensemble model. This integration
is depicted in the variable importance plot of Figure \ref{fig:varimp}
(b), which places a strong emphasis on the prediction results from
collaborators 1 to 3 (i.e., \textquotedbl col1\_proba\textquotedbl ,
\textquotedbl col2\_proba\textquotedbl , and \textquotedbl col3\_proba\textquotedbl{}
on the x-axis). This contrasts with the previous simulation where
Collaborator 1 was learning independently, as shown in Figure \ref{fig:varimp}
(a).

Consider this, even though Collaborator 1 achieves high performance
through collaboration, some of the features contributing to this success
originate from Collaborators 2, 3, and 4. As Collaborator 1 cannot
independently regenerate these features (i.e., prediction results),
its prediction power will be severely impacted upon leaving the collaboration,
irrespective of the number of labels it possesses in the new domain.

Comparing Figure \ref{fig:auc-score}(a), which depicts independent
learning, with Figure \ref{fig:auc-score}(b), illustrating four collaborators
engaged in cooperative active learning, the advantages of collaborative
active learning become evident.  It can be argued that Collaborator
4, possessing a base model with high AUC score need not engage in
collaboration. However, it is important to note that collaborators
are only privy to newly acquired labels and remain unaware of their
model's performance in a new domain. Therefore, as discussed in Assumption
\ref{enu:improvement-ensemble-learning} of Section \ref{subsec:Retrain-Models},
Collaborator 4 will preferably employ collaborative active learning
to minimize the risk associated with exploring a new domain. 

\section{Conclusion\label{sec:Conclusion}}

This paper presents a collaborative active learning framework that
aims to leverage the collective machine learning capabilities in C2AL
environment, where data and models are not shared among the participants.
Our results unequivocally show that collaborative active learning
significantly outperforms independent active learning strategies.
This underscores the synergistic potential of collaborative efforts
in active learning environments.

\bibliographystyle{IEEEtran}
\bibliography{bibliography}

\begin{thebibliography}{10}
\providecommand{\url}[1]{#1}
\csname url@samestyle\endcsname
\providecommand{\newblock}{\relax}
\providecommand{\bibinfo}[2]{#2}
\providecommand{\BIBentrySTDinterwordspacing}{\spaceskip=0pt\relax}
\providecommand{\BIBentryALTinterwordstretchfactor}{4}
\providecommand{\BIBentryALTinterwordspacing}{\spaceskip=\fontdimen2\font plus
\BIBentryALTinterwordstretchfactor\fontdimen3\font minus
  \fontdimen4\font\relax}
\providecommand{\BIBforeignlanguage}[2]{{%
\expandafter\ifx\csname l@#1\endcsname\relax
\typeout{** WARNING: IEEEtran.bst: No hyphenation pattern has been}%
\typeout{** loaded for the language `#1'. Using the pattern for}%
\typeout{** the default language instead.}%
\else
\language=\csname l@#1\endcsname
\fi
#2}}
\providecommand{\BIBdecl}{\relax}
\BIBdecl

\bibitem{goldmansachs2019apple}
\BIBentryALTinterwordspacing
G.~Sachs, ``Goldman sachs partners with apple on a game-changing credit card,''
  \emph{Goldman Sachs | Commemorates 150 Year History}, 2019. [Online].
  Available:
  \url{https://www.goldmansachs.com/our-firm/history/moments/2019-apple-card.html}
\BIBentrySTDinterwordspacing

\bibitem{kapron2021malaysian}
\BIBentryALTinterwordspacing
Z.~Kapron, ``Malaysian digital banks: 29 applicants for a maximum of 5
  licenses,'' \emph{Forbes}, Jul 2021. [Online]. Available:
  \url{https://www.forbes.com/sites/zennonkapron/2021/07/19/malaysian-digital-banks-29-applicants-for-a-maximum-of-5-licenses/}
\BIBentrySTDinterwordspacing

\bibitem{lewis1994heterogeneous}
D.~D. Lewis and J.~Catlett, ``Heterogeneous uncertainty sampling for supervised
  learning,'' in \emph{Machine learning proceedings 1994}.\hskip 1em plus 0.5em
  minus 0.4em\relax Elsevier, 1994, pp. 148--156.

\bibitem{freund1997selective}
Y.~Freund, H.~S. Seung, E.~Shamir, and N.~Tishby, ``Selective sampling using
  the query by committee algorithm,'' \emph{Machine learning}, vol.~28, pp.
  133--168, 1997.

\bibitem{Chong2023}
Z.-K. Chong, H.~Ohsaki, and B.-M. Goi, ``Improving uncertainty sampling with
  bell curve weight function,'' \emph{International Journal of Applied Physics
  and Mathematics}, vol.~13, no.~4, pp. 44--52, 2023.

\bibitem{tharwat2023survey}
A.~Tharwat and W.~Schenck, ``A survey on active learning: State-of-the-art,
  practical challenges and research directions,'' \emph{Mathematics}, vol.~11,
  no.~4, p. 820, 2023.

\bibitem{GoogleFederatedLearning}
\BIBentryALTinterwordspacing
Google. (2017) Federated learning: Collaborative machine learning without
  centralized training data. [Online]. Available:
  \url{https://blog.research.google/2017/04/federated-learning-collaborative.html}
\BIBentrySTDinterwordspacing

\bibitem{wang2023unlocking}
H.~Wang, X.~Liu, J.~Niu, S.~Tang, and J.~Shen, ``Unlocking the potential of
  federated learning for deeper models,'' \emph{arXiv preprint
  arXiv:2306.02701}, 2023.

\bibitem{wen2023survey}
J.~Wen, Z.~Zhang, Y.~Lan, Z.~Cui, J.~Cai, and W.~Zhang, ``A survey on federated
  learning: challenges and applications,'' \emph{International Journal of
  Machine Learning and Cybernetics}, vol.~14, no.~2, pp. 513--535, 2023.

\bibitem{warnat2021swarm}
S.~Warnat-Herresthal, H.~Schultze, K.~L. Shastry, S.~Manamohan, S.~Mukherjee,
  V.~Garg, R.~Sarveswara, K.~H{\"a}ndler, P.~Pickkers, N.~A. Aziz
  \emph{et~al.}, ``Swarm learning for decentralized and confidential clinical
  machine learning,'' \emph{Nature}, vol. 594, no. 7862, pp. 265--270, 2021.

\bibitem{han2022demystifying}
J.~Han, Y.~Ma, and Y.~Han, ``Demystifying swarm learning: A new paradigm of
  blockchain-based decentralized federated learning,'' \emph{arXiv preprint
  arXiv:2201.05286}, 2022.

\bibitem{saldanha2022swarm}
O.~L. Saldanha, P.~Quirke, N.~P. West, J.~A. James, M.~B. Loughrey, H.~I.
  Grabsch, M.~Salto-Tellez, E.~Alwers, D.~Cifci, N.~Ghaffari~Laleh
  \emph{et~al.}, ``Swarm learning for decentralized artificial intelligence in
  cancer histopathology,'' \emph{Nature Medicine}, vol.~28, no.~6, pp.
  1232--1239, 2022.

\bibitem{li2022can}
Z.~Li, F.~Mao, and C.~Wu, ``Can we share models if sharing data is not an
  option?'' \emph{Patterns}, vol.~3, no.~11, 2022.

\bibitem{dietterich2000ensemble}
T.~G. Dietterich, ``Ensemble methods in machine learning,'' in
  \emph{International workshop on multiple classifier systems}.\hskip 1em plus
  0.5em minus 0.4em\relax Springer, 2000, pp. 1--15.

\bibitem{akano2022assessment}
T.~T. Akano and C.~C. James, ``An assessment of ensemble learning approaches
  and single-based machine learning algorithms for the characterization of
  undersaturated oil viscosity,'' \emph{Beni-Suef University Journal of Basic
  and Applied Sciences}, vol.~11, no.~1, pp. 1--18, 2022.

\bibitem{naderalvojoud2023improving}
B.~Naderalvojoud and T.~Hernandez-Boussard, ``Improving machine learning with
  ensemble learning on observational healthcare data,'' in \emph{AMIA Annual
  Symposium Proceedings}, vol. 2023.\hskip 1em plus 0.5em minus 0.4em\relax
  American Medical Informatics Association, 2023, p. 521.

\bibitem{li2020comparative}
Y.~Li and W.~Chen, ``A comparative performance assessment of ensemble learning
  for credit scoring,'' \emph{Mathematics}, vol.~8, no.~10, p. 1756, 2020.

\bibitem{scikit-learn}
F.~Pedregosa, G.~Varoquaux, A.~Gramfort, V.~Michel, B.~Thirion, O.~Grisel,
  M.~Blondel, P.~Prettenhofer, R.~Weiss, V.~Dubourg, J.~Vanderplas, A.~Passos,
  D.~Cournapeau, M.~Brucher, M.~Perrot, and E.~Duchesnay, ``Scikit-learn:
  Machine learning in {P}ython,'' \emph{Journal of Machine Learning Research},
  vol.~12, pp. 2825--2830, 2011.

\end{thebibliography}

\end{document}